\documentclass[10pt,twocolumn,letterpaper]{article}

\usepackage[pagenumbers]{wacv} % To force page numbers, e.g. for an arXiv version

\usepackage{graphicx}
\usepackage{amsmath}
\usepackage{amssymb}
\usepackage{booktabs}

\usepackage[pagebackref,breaklinks,colorlinks]{hyperref}

\usepackage[capitalize]{cleveref}
\crefname{section}{Sec.}{Secs.}
\Crefname{section}{Section}{Sections}
\Crefname{table}{Table}{Tables}
\crefname{table}{Tab.}{Tabs.}

\def\wacvPaperID{2424} % *** Enter the WACV Paper ID here
\def\confName{WACV}
\def\confYear{2024}

\begin{document}

\title{Uncertainty Estimation in Instance Segmentation with Star-convex Shapes}

\author{Qasim M. K. Siddiqui, Sebastian Starke, and Peter Steinbach\\
Hemholtz-Zentrum Dresden-Rossendorf\\
Dresden, Germany\\
{\tt\small q.siddiqui@hzdr.de, s.starke@hzdr.de, p.steinbach@hzdr.de}
}
\maketitle

%%%%%%%%% ABSTRACT
\begin{abstract}
   Instance segmentation has witnessed promising advancements through deep neural network-based algorithms. However, these models often exhibit incorrect predictions with unwarranted confidence levels. Consequently, evaluating prediction uncertainty becomes critical for informed decision-making. Existing methods primarily focus on quantifying uncertainty in classification or regression tasks, lacking emphasis on instance segmentation. Our research addresses the challenge of estimating spatial certainty associated with the location of instances with star-convex shapes. Two distinct clustering approaches are evaluated which compute spatial and fractional certainty per instance employing samples by the Monte-Carlo Dropout or Deep Ensemble technique. Our study demonstrates that combining spatial and fractional certainty scores yields improved calibrated estimation over individual certainty scores. Notably, our experimental results show that the Deep Ensemble technique alongside our novel radial clustering approach proves to be an effective strategy. Our findings emphasize the significance of evaluating the calibration of estimated certainties for model reliability and decision-making.
\end{abstract}

%%%%%%%%% BODY TEXT
\section{Introduction}
\label{sec:intro}

In the past decade, deep neural networks have made significant advancements and have become prevalent in the field of computer vision, achieving impressive state-of-the-art performance and even competing with human-level results in supervised learning tasks~\cite{ImageNet, reinf, spchrec, deeptrac}. However, these achievements are primarily observed in closed-set conditions, where the testing data exhibits overlapping characteristics with the training data. In contrast, a noticeable decline in performance occurs in open-set conditions~\cite{open_set_1, open_set_2}, where the testing data possess characteristics not present in the training data. In such scenarios, neural network models often make wrong predictions with high confidence~\cite{open_set_3}, which raises critical concerns about the safety and reliability of deploying these models, particularly in applications where perception errors can have severe consequences~\cite{miller_4}. To address this challenge, one promising approach is to explore the model's epistemic uncertainty, which arises due to a lack of data~\cite{ae_matter}. High epistemic uncertainty in predictions can potentially indicate open-set errors~\cite{drop_obj_01, drop_obj_02}, enabling models to identify and handle such detections appropriately.

Bayesian Neural Networks~\cite{bprop, bnn_phd, bnn} offer a means to estimate epistemic uncertainty. However, their practical application is limited due to the higher computational cost and training complexity involved. To address these challenges, Monte-Carlo Dropout was introduced by Gal \etal in 2015~\cite{mcd} as a computationally feasible approximation to Bayesian neural networks, providing uncertainty estimates for a model's confidence scores.

While Monte-Carlo Dropout offers a feasible technique for estimating uncertainty, it often requires extensive hyperparameter tuning to obtain well-calibrated predictive uncertainty estimates~\cite{drop_rate}. In response, Deep Ensemble techniques, a non-Bayesian solution~\cite{deep_ens}, were introduced, which yield well-calibrated predictive uncertainty estimates with minimal hyperparameter tuning at the cost of requiring multiple model training.

Recently, Monte-Carlo Dropout and Deep Ensemble techniques have shown promising results in uncertainty estimation for image classification and regression tasks~\cite{mcd, deep_ens}. However, their application to instance segmentation, which involves localizing and classifying multiple objects within a scene, remains relatively underexplored.

Our research aims to bridge this gap by applying Monte-Carlo Dropout and Deep Ensemble techniques to the specific domain of instance segmentation using the StarDist model~\cite{stardist_01, stardist_02}. The effectiveness of these techniques in estimating certainty in instance segmentation was verified through extensive experiments on three different datasets. Specifically, our research evaluates and compares the effectiveness of both Monte-Carlo Dropout and Deep Ensemble techniques for the StarDist model. Furthermore, we investigate the effect of dropout rates and the location of the dropout layer within the StarDist model for Monte-Carlo Dropout certainty estimation in instance segmentation.

By addressing these research gaps, this study enhances the reliability and robustness of deep neural networks by providing well-calibrated predictive certainty estimates for instance segmentation. These advancements enable more informed decision-making and improve the performance of deep neural networks in practical applications of instance segmentation with the StarDist model.

%-------------------------------------------------------------------------
\section{Related Work}
\label{sec:related_work}

Uncertainty estimation in deep neural networks has been a subject of extensive research, focusing on two types of uncertainty: aleatoric uncertainty and epistemic uncertainty~\cite{ae_matter}.

Aleatoric uncertainty also known as data uncertainty, captures the inherent noise present in the data itself. It represents the statistical or sensory noise that cannot be reduced even with an increase in the amount of collected data~\cite{what_unc}. Aleatoric uncertainty can be directly learned using neural networks by applying a distribution over the model's output, enabling the model to capture varying levels of uncertainty for different inputs~\cite{what_unc}. 

Epistemic uncertainty also referred to as model uncertainty, arises from parameter ambiguity and limited knowledge about the model. It reflects the uncertainty in the model's predictions and requires additional techniques to estimate. One approach is to place a prior distribution over the model's parameters and analyze the variability of this distribution given the available data~\cite{bnn, bnn_phd, par_dist}. Estimating epistemic uncertainty is particularly crucial for safety-critical systems and models trained with small datasets, as it helps identify situations that lie beyond the model's training data.

Estimating epistemic uncertainty in deep neural networks presents challenges not encountered in traditional machine learning algorithms~\cite{survey_active}. The inherent inability of deep neural networks to accurately quantify uncertainty has led to investigations into alternative approaches. Bayesian Neural Networks offer a means of predicting output uncertainty~\cite{bnn}, but their computational demands often render them impractical~\cite{dropout, drop_obj_01}. 

To address this, Gal \etal~\cite{mcd} proposed using dropout in the last layers of deep learning models during test time, with multiple forward passes, to approximate Bayesian inference over network parameters. This dropout-based sampling approach better captures input-specific uncertainty within the model. It has been successfully applied to tasks such as active learning for image categorization~\cite{drop_active}, melanoma identification~\cite{unc_melanoma}, and object recognition using LiDAR data~\cite{unc_lidar}.

While dropout-based sampling initially focused on deep neural network classification tasks, Miller \etal~\cite{drop_obj_01} extended the concept to Single Shot Detection (SSD) for object detection~\cite{ssd}. In this more complex task, each forward pass generates multiple object detections that need to be matched and merged. By clustering detections based on spatial and semantic similarity, ambiguous detections are rejected, leading to improved object detection performance in both closed and open-set scenarios. Miller \etal~\cite{drop_obj_02} further explored this approach, evaluating alternative strategies for merging detections while incorporating dropout-based sampling in object detection.

Morrison \etal~\cite{amazone} built upon these foundations to address pixel-wise masked instance segmentation. They adapted the technique proposed in~\cite{drop_obj_01} to perform probabilistic instance segmentation. Leveraging the Mask-RCNN network~\cite{m_rcnn}, they employed dropout-based sampling during inference, following the principles established in Srivastava \etal~\cite{dropout} and Miller \etal~\cite{drop_obj_01}. This methodology resulted in a well-calibrated uncertainty estimation.

Overall, these studies highlight the significance of dropout-based sampling techniques in capturing and quantifying uncertainty in deep neural networks, spanning various tasks ranging from classification to object detection and instance segmentation.

Our study extends the work by Morrison \etal~\cite{amazone} to the StarDist model, which is a novel approach to instance segmentation, addressing the limitations of existing methods like Mask-RCNN in handling crowded instances~\cite{stardist_01, stardist_02}. 

The StarDist model utilizes a U-Net architecture as its building block, which is well-suited for image segmentation tasks and has demonstrated state-of-the-art performance in biomedical image segmentation~\cite{unet}.

\subsection{Revisiting StarDist}

\begin{figure}[t]
  \centering
  \includegraphics[width=0.8\linewidth]{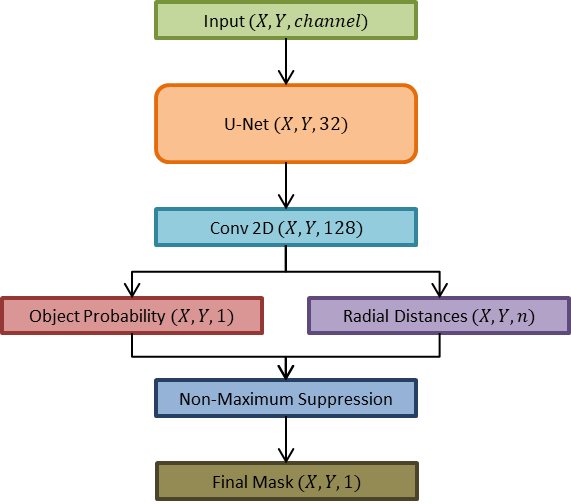}

   \caption{The fundamental structure of the StarDist model involves an input image with dimensions $(X, Y, channel)$. Within this model, a U-net is employed, featuring two output layers: Object Probability $d_{x,y} \in D$ and Radial Distances $\{r_{x, y}^{i}\}_{i = 1}^{n} \in R$, both of which constitute scalar fields matching the input image size. The process concludes with the derivation of the Final Mask after applying non-maximum suppression to both the Object Probability $(D)$ and Radial Distances $(R)$.
   }
   \label{fig:stardist}
\end{figure}

The StarDist model proposes the localization of convex-shaped instances using star-convex polygons, yielding satisfactory outcomes for densely populated scenarios~\cite{stardist_01}. The fundamental component of the StarDist model is depicted in Figure~\ref{fig:stardist}.

The StarDist model, akin to object detection approaches~\cite{obj_1, ssd, obj_3}, employs a star-convex polygon prediction for each pixel. Specifically, for every pixel $(x, y)$ within the scalar field of the input image size $(X, Y, channel)$, we predict the radial distances $\{r_{x, y}^{i}\}_{i = 1}^{n}$ to the object boundary. These predictions are made along a predefined set of $n$ radial directions with equidistant angles. The model independently predicts whether each pixel is part of an object, focusing on polygon proposals from pixels with sufficiently high object probability $d_{x,y}$. By considering these dense polygon candidates and their associated object probabilities, we employ non-maximum suppression to derive the final collection of polygons. Each polygon represents a distinct object instance.

%-------------------------------------------------------------------------

\section{Methods}
\label{sec:methods}

In our research, the StarDist model is adapted to incorporate the Deep Ensemble~\cite{deep_ens} and Monte-Carlo Dropout~\cite{mcd} techniques during inference, enabling probabilistic instance segmentation similar to prior works of Miller \etal~\cite{drop_obj_01} and Morrison \etal~\cite{amazone}. These sampling techniques utilize multiple output samples obtained from $F$ forward passes to assess the certainty associated with the model's predictions.

\begin{figure}[t]
  \centering
  \includegraphics[width=0.95\linewidth]{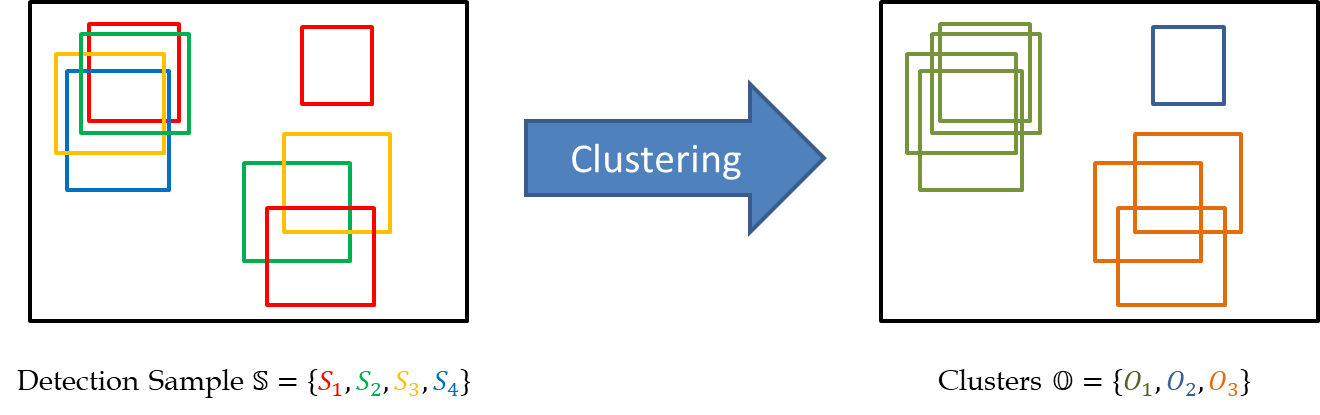}

   \caption{An illustration of image clustering based on predictions obtained from $F = 4$ forward passes, with $\mid S_1 \mid = 3$, $\mid S_2 \mid = 2$, $\mid S_3 \mid = 2$, and $\mid S_4 \mid = 1$. The aim is to group predicted instances that correspond to the same object into distinct clusters $\mathbb{O}$. In this example, $\mid O_1 \mid = 4$, $\mid O_2 \mid = 1$, and $\mid O_3 \mid = 3$, where each cluster $O_m \mid m \in \{1, 2, 3\}$ represents an instance.
   }
   \label{fig:fig_cluster}
\end{figure}

While sampling techniques can be employed for estimating epistemic certainty in tasks like classification or semantic segmentation by simply averaging the samples, instance segmentation poses unique challenges. In instance segmentation, accurate association and clustering of different detection samples are required to identify multiple instances within an image, as depicted in Figure~\ref{fig:fig_cluster}. To account for spatial certainty, similar to the approaches in Miller \etal~\cite{drop_obj_01} and Miller \etal~\cite{drop_obj_02}, certainty estimates are obtained by integrating instances from successive forward passes, as depicted in Figure~\ref{fig:fig_cluster}.

The StarDist model comprises two sets of outputs: $(a)$ the final mask and $(b)$ pixel-wise object probability predictions and radial distance, each of these output sets can be utilized individually to calculate the certainty associated with the model's predictions. In the subsequent sections, we present approaches that leverage these output sets to quantify the model's certainty.

\subsection{Pixel Approach}
\label{sec:pixel_approach}

In the first approach for calculating certainty, we utilize the final mask containing the collection of polygons obtained after non-maximum suppression, which have the same dimensions as the input image. Each pixel in the final mask is assigned a $positive$ integer value if it is within a polygon representing an instance or $zero$ if it belongs to the background.

\subsubsection{Clustering Technique}
To estimate certainty with the final mask from the StarDist model, we adopt a clustering technique inspired by prior work on object detection and instance segmentation~\cite{drop_obj_01, drop_obj_02, amazone}.

In this approach, each forward pass of input through the StarDist model generates a set of predicted instances $S = \{P_1, P_2, ..., P_{K}\}$, where $K$ represents the number of predicted instances, which may vary across different forward passes. In the case of binary class prediction, $P_k \mid k \in \{1, 2, ..., K\}$ represents the pixel-wise mask of the instance for a given input image. By performing $F$ forward passes of the input image through the model, we obtain a set of samples $\mathbb{S} = \{S_1, S_2, ..., S_F\}$, where each $S_f = \{P_{1_f}, P_{2_f}, ..., P_{k_{f}}\}$ contains a set of predictions, where $k_{f} \mid f \in \{1, 2, ..., F\}$ represents the number of predicted instances, which may vary across different forward passes.

Based on their spatial affinity, the predictions from the set of samples obtained through all forward passes $\mathbb{S}$ are grouped into individual clusters $\mathbb{O} = \{O_1, O_2, ..., O_M\}$. Ideally, each cluster $O_m \mid m \in \{1,2, ..., M\} $ should represent a single instance within the image.% This clustering process is illustrated in Figure~\ref{fig_cluster}.

To perform the clustering of instances, we employ the Basic Sequential Algorithmic Scheme (BSAS)~\cite{drop_obj_02}, where predictions $P_{k_f}$ are sequentially assigned to clusters $O_m$ if their mask Intersection-over-Union $(IoU)$\footnote{See Supplementary Material for Intersection-over-Union $(IoU)$ calculation.} exceeds a threshold value $\theta_{IoU}$. If the $IoU$ between a prediction and every prediction in an existing cluster is above the threshold $\theta_{IoU}$, the prediction is added to that cluster. If no existing cluster matches the prediction, a new cluster is created. The $IoU$ threshold $\theta_{IoU}$ is subject to optimization. However we fix $\theta_{IoU} = 0.5$~\cite{amazone}. The clustering algorithm in the Pixel Approach is given in the Supplementary Material.

\subsubsection{Visualization of Prediction and Uncertainty}
\begin{figure}[t]
  \centering
  \includegraphics[width=0.95\linewidth]{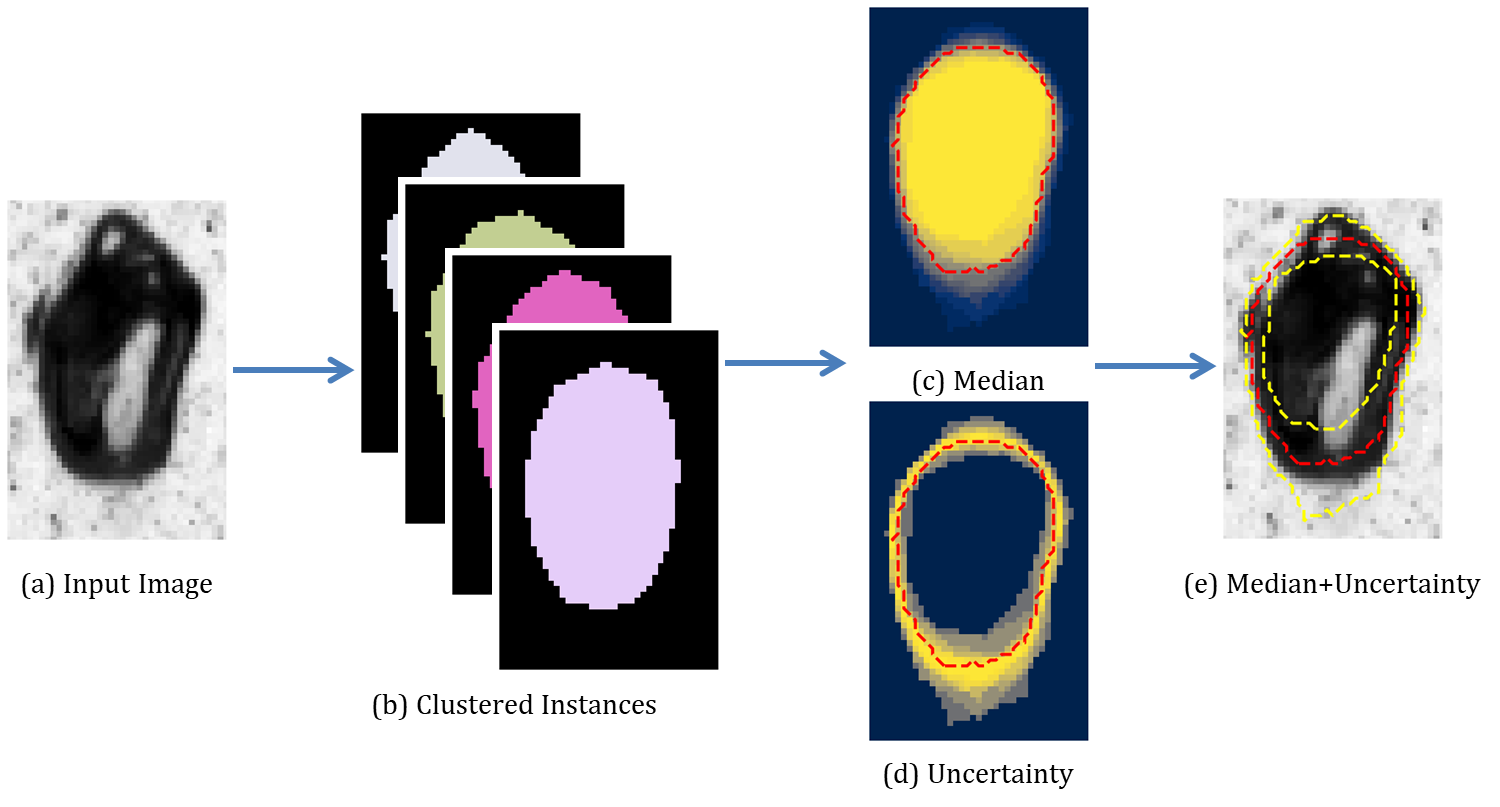}

   \caption{An illustration of the process of visualizing the uncertainty in the Pixel Approach. (a) Input Image, (b) Clustered instance ($O_m$), (c) Mean calculated for each pixel, (d) Standard deviation calculated for each pixel, and (e) Red polygons illustrate the median cluster prediction $\overline{P}_m$, and the region between the two yellow polygons indicates the uncertainty of the instance.
   }
   \label{fig:vis_pixel}
\end{figure}

To visualize the median cluster prediction $\overline{P}_m$ for cluster $O_m$, we calculate the median across all pixels $(x_o,y_o) \mid o=\{1, 2, ..., \mid O_m \mid\}$. The pixel value of the median cluster prediction $\overline{P}_m$ is set to $integer$ if it corresponds to an instance and $zero$ otherwise. In Figure \ref{fig:vis_pixel}, the red polygons illustrate the median cluster prediction $\overline{P}_m$.

To visualize uncertainty, we determine the mean and the standard deviation values across all pixels $(x_o,y_o) \mid o=\{1, 2, ..., \mid O_m \mid\}$ within cluster $O_m$. A pixel with a low standard deviation signifies low uncertainty that it belongs to the instance while increasing standard deviation values reflect increasing uncertainty. In Figure \ref{fig:vis_pixel}, the contour of the standard deviation values is represented by the yellow line, conveying its associated uncertainty level. Pixels enclosed by the inner yellow polygons indicate a low degree of uncertainty in belonging to a specific instance, while those outside the outer yellow polygons indicate a low degree of uncertainty in not belonging to a specific instance in question. The region between the two yellow polygons indicates the uncertainty of a specific instance.

\subsection{Radial Approach}
\label{sec:radial_approach}

In this approach for calculating certainty, we leverage the output structure of the StarDist model, consisting of pixel-wise object probability predictions and radial distance, which are obtained before non-maximum suppression. We will be using the term $Dense Output$ ($G = \{D, R\}$) to refer to the scalar field of object probabilities ($d_{x,y} \in D$) and radial distances ($\{r_{x,y}^{n}\}_{i=1}^{n} \in R$) for improved clarity and conciseness.

The clustering approach of instance masks based on $IoU$ from the Pixel Approach is replaced by identifying the centers of the instances from the sample mean of the set of $Dense Output$, and clustering instances whose object probability ($d_{x,y} \in D$) is above a predefined threshold value ($\theta_d$).
\begin{figure}[t]
  \centering
  \includegraphics[width=0.95\linewidth]{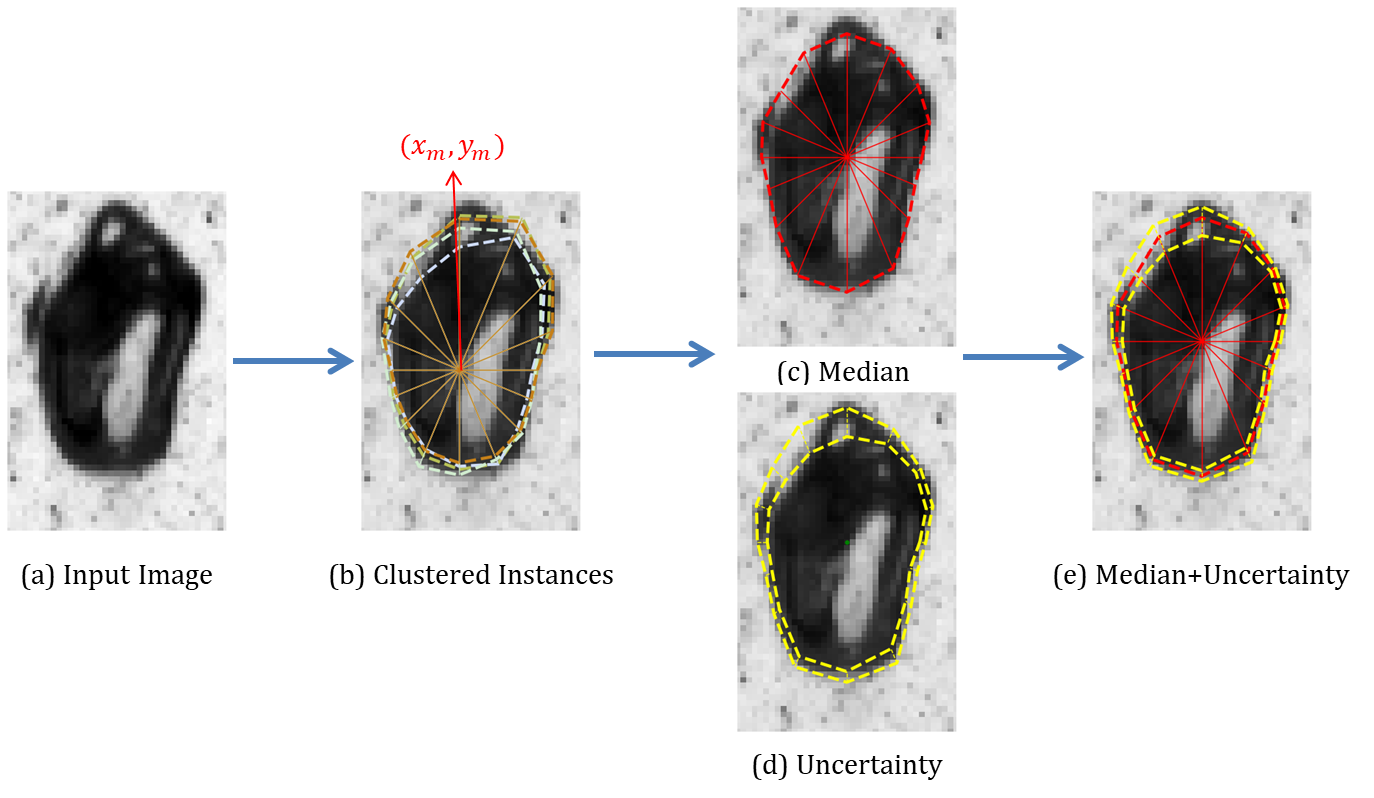}

   \caption{An illustration of the process of visualizing the uncertainty in the Pixel Approach. (a) Input Image, (b) Clustered instance ($O_m$) for polygon center $(x_m, y_m)$, (c) Median calculated across all the radial distances, (d) $2.5^{th}$ and $97.5^{th}$ percentile across all the radial distances, and (e) Red polygons illustrate the median cluster prediction $\overline{P}_m$, the inner yellow polygons are indicated by the $2.5^{th}$ percentile of the radial distances, while the outer yellow polygons are indicated by the $97.5^{th}$ percentile of the radial distances.
   }
   \label{fig:vis_radial}
\end{figure}

\subsubsection{Clustering Technique}

In this approach, each forward pass of an input through the StarDist model generates the $Dense Output$ $G = \{D, R\}$. By performing $F$ forward passes of the input image through the model, we obtain a set of $Dense Output$ samples $\mathbb{G} = \{G_1, G_2, ..., G_F\}$, where $G_f = \{D_f, R_f\} \mid f \in \{1, 2, ..., F\}$.

To perform the clustering of instances, we identify the center for each instance by taking the mean of the set $\mathbb{G}$ to parameterize it into a single $Dense Output$ $\mu_{G} = \{\mu_{D}, \mu_{R}\}$, where $\mu_{D}$ and $\mu_{R}$ represent the mean object probability and mean radial distance across the samples $\mathbb{G}$. Non-maximum suppression is applied to $\mu_{G}$ to obtain the set of polygon centers with the highest object probability, $\mathbb{C} = \{(x_1, y_1), (x_2, y_2), .., (x_M, y_M)\}$. 

The combination of the polygon center $(x_m, y_m) \mid m \in \{1, 2, ..., M\}$ and the radial distances at the corresponding polygon centers $\{\{r_{x_m, y_m}^{n}\}_{i=1}^{n}\}_f \in R_f \mid f \in \{1, 2, ..., F\}$ a set of predictions instance are generated $S = \{P_1, P_2, ..., P_M\}$ for each forward pass. By performing $F$ forward passes of the input image through the model, we obtain a set of samples $\mathbb{S} = \{S_1, S_2, ..., S_F\}$, where each $S_f = \{P_{1_f}, P_{2_f}, ..., P_{M_{f}}\}$ contains a set of predictions, where $M_{f} \mid f \in \{1, 2, ..., F\}$ represents the number of predicted instances, which in this case is fixed at $M$ across different forward passes.

The predictions from the set of samples obtained through all forward pass $\mathbb{S}$ are grouped into individual clusters $\mathbb{O} = \{O_1, O_2, ..., O_M\}$. Ideally, each cluster  $O_m \mid m \in \{1,2, ..., M\}$ should represent a single object within the image with $(x_m, y_m) \mid m \in \{1, 2, ..., M\}$ as individual polygon center. 

The predictions $P_{m_f}$ are sequentially assigned to the cluster $O_m$ if the object probability $\{d_{x_m, y_m}\}_f$ exceeds a threshold value $\theta_d$. The object probability threshold $\theta_{d}$ is subject to optimization. However, for consistency with the Pixel Approach \ref{sec:pixel_approach}, we set $\theta_{d} = 0.5$. The clustering algorithm in the Radial Approach is given in the Supplementary Material.

\subsubsection{Visualization of Prediction and Uncertainty}
To visualize the median cluster prediction $\overline{P}_m$ for cluster $O_m$, we calculate the median across all the radial distances $\{\{r_{x_m, y_m}^{n}\}_{i=1}^{n}\}_o \mid o=\{1, 2, ..., \mid O_m \mid\}$. In Figure \ref{fig:vis_radial}, the red polygons illustrate the median cluster prediction $\overline{P}_m$.

We visualize uncertainty, by determining the $2.5^{th}$ and $97.5^{th}$ percentile across all the radial distances $\{\{r_{x_m, y_m}^{n}\}_{i=1}^{n}\}_o \mid o=\{1, 2, ..., \mid O_m \mid\}$. In Figure \ref{fig:vis_radial}, the inner yellow polygons are indicated by the $2.5^{th}$ percentile of the radial distances, while the outer yellow polygons are indicated by the $97.5^{th}$ percentile of the radial distances for the specific instance. The region between the two yellow polygons indicates the spatial uncertainty of a specific instance.

\subsection{Certainty Quantification}

Once all predictions $P_{{\cdot}_f}$ are grouped into clusters $\mathbb{O}$, we can quantify the certainty of our model's predictions using two scores for each cluster $O_m \mid m \in \{1, 2, ..., M\}$:

\textbf{(i) Spatial Certainty:} This score informs us about the model's confidence in the location of each instance. The spatial certainty for each cluster $O_m$ is computed by averaging the intersection-over-union ($IoU$) between the median cluster prediction $\overline{P}_m$ and the prediction $P_j$ of each of the $\mid O_m \mid$ predictions within the cluster $O_m$. The spatial certainty ranges from $0$ to $1$, where $1$ indicates high certainty and $0$ indicates low certainty.
\begin{equation}
        \label{spatial_unc}
        c_{spl}(O_m) = \frac{1}{\mid O_m \mid} \sum_{j=1}^{\mid 0_m \mid} IoU(P_j, \overline{P}_m)
\end{equation}

\textbf{(ii) Fractional Certainty:} This score represents the model's confidence in detecting an instance across multiple forward passes. As not every forward pass may detect all instances, the fractional certainty for each cluster $O_m$ is computed as the fraction of forward passes in which the instance is detected. Similarly, the fractional certainty ranges from $0$ to $1$, where $1$ indicates the instance was predicted in all forward passes and $0$ indicates the instance was not detected in any forward pass.
\begin{equation}
        \label{frac_unc}
        c_{frac}(O_m) = \frac{\mid O_m \mid}{N} 
\end{equation}

\textbf{(iii) Hybrid Certainty:} It has been observed by Morrison \etal~\cite{amazone} that combining the two certainty scores results in well-calibrated certainty estimates compared to using them separately.
\begin{equation}
        \label{hybrid_unc_pixel}
        c_{hyb}(O_m) = c_{spl}(O_m) \cdot c_{frac}(O_m)
\end{equation}

\subsection{Evaluating Certainty}

Calibration serves as a valuable metric to evaluate the estimated certainty of the model.
Model calibration refers to the accuracy and precision of the certainty score in indicating when the model is likely to make errors. Calibration is crucial for interpretability and building trust in users who are consuming model prediction. A well-calibrated model provides reliable certainty scores that align with its predictive expected accuracy~\cite{calibration}.
\begin{figure*}[t]
  \centering
  \subfloat[][]{
    \includegraphics[width=0.29\textwidth]{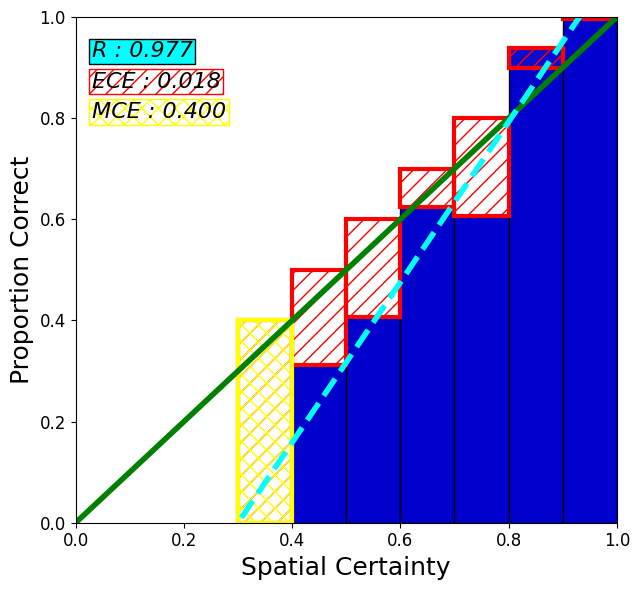}
    \label{fig:cert_quality_spl}}
    \qquad
    \subfloat[][]{
    \includegraphics[width=0.29\textwidth]{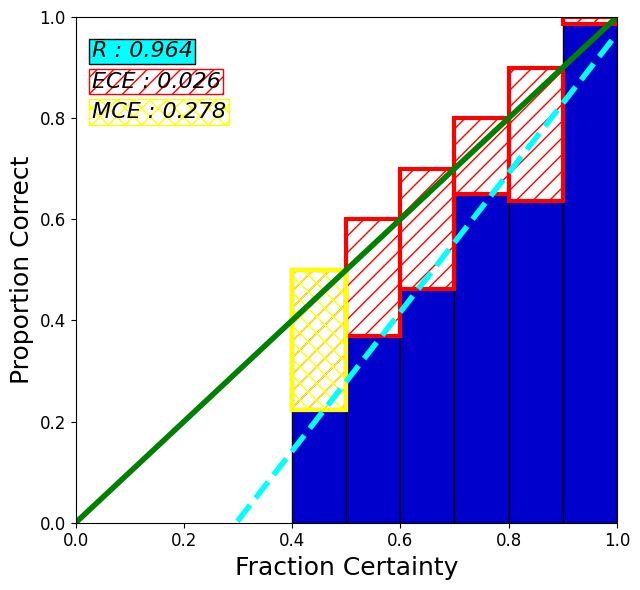}
    \label{fig:cert_quality_frac}}
    \qquad
    \subfloat[][]{
    \includegraphics[width=0.29\textwidth]{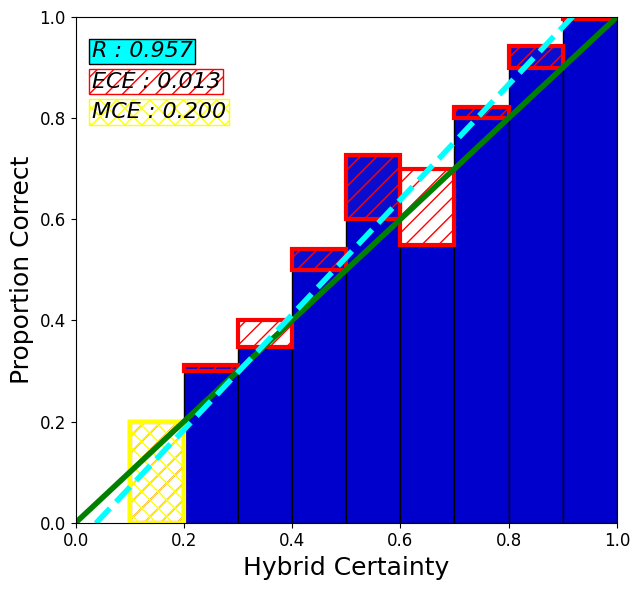}
    \label{fig:cert_quality_hyb}}
  \caption{Calibration diagrams depicting the estimation of certainty scores for the Bubble dataset are presented, employing a bin interval of $B=10$. Panels (a), (b), and (c) show spatial certainty $(c_{spl})$, fractional certainty $(c_{frac})$, and hybrid certainty $(c_{hyb})$ scores, respectively. These scores are calculated using the Radial Approach and Monte-Carlo Dropout with a dropout rate of $d_{rate} = 0.8$, and $F = 20$ forward passes. Notably, the hybrid certainty scores $(c_{hyb})$ demonstrate superior calibration compared to individual certainty scores across three calibration error metrics: Pearson Correlation Coefficient (R), Expected Calibration Error (ECE), and Maximum Calibration Error (MCE).
  }
  \label{fig:cert_quality}
\end{figure*}

\subsubsection{Calibration Diagram}
A calibration diagram visualizes the expected sample accuracy as a function of the certainty score. To create the diagram, predictions are divided into $B$ interval bins of size $1/B$. The expected accuracy and average certainty score within each bin are computed. If a model is perfectly calibrated, the expected accuracy and certainty score should be equal for each bin~\cite{cali_visual_01, cali_visual_02}. This is visualized in Figure \ref{fig:cert_quality}, which compares the three certainty quantification scores for the StarDist model with the Radial Approach and Monte-Carlo Dropout $d_{rate} = 0.8$, with $F = 20$.

\subsubsection{Calibration Error}

Scalar calibration summaries are practical for evaluating calibration diagrams. The commonly used metrics are the Pearson correlation coefficient, the Expected Calibration Error, and the Maximum Calibration Error, which can be derived from the calibration diagrams.

\textbf{Pearson Correlation Coefficient (Pearson's R)} measures the linear correlation between two sets of data. In this context, it quantifies the correlation between the identity function and the bin scores. The coefficient ranges from -1 to 1, with 0 indicating no correlation, negative values indicating negative correlation, and positive values indicating positive correlation~\cite{prob_thr}.

\textbf{Expected Calibration Error (ECE)} approximates the expected difference between the certainty score and expected accuracy. It involves dividing predictions into evenly spaced bins (similar to Calibration Diagram Figure \ref{fig:cert_quality}) and averaging the accuracy difference within each bin. Expected Calibration Error provides a measure of miscalibration, with a perfectly calibrated model having an Expected Calibration Error of zero~\cite{ece}.

\textbf{Maximum Calibration Error (MCE)} summarizes miscalibration by measuring the maximum difference between the certainty score and expected accuracy. This metric is useful in high-risk applications where minimizing the worst-case difference is crucial. Maximum Calibration Error is defined as the maximum absolute difference across all bins. Like Expected Calibration Error, a perfectly calibrated model will have a Maximum Calibration Error of zero~\cite{ece}.
%-------------------------------------------------------------------------
\section{Experiments and Results}
\subsection{Datasets}
\begin{figure}[t]
  \centering
    \subfloat[][Bubble Image]{
    \includegraphics[width=0.15\textwidth]{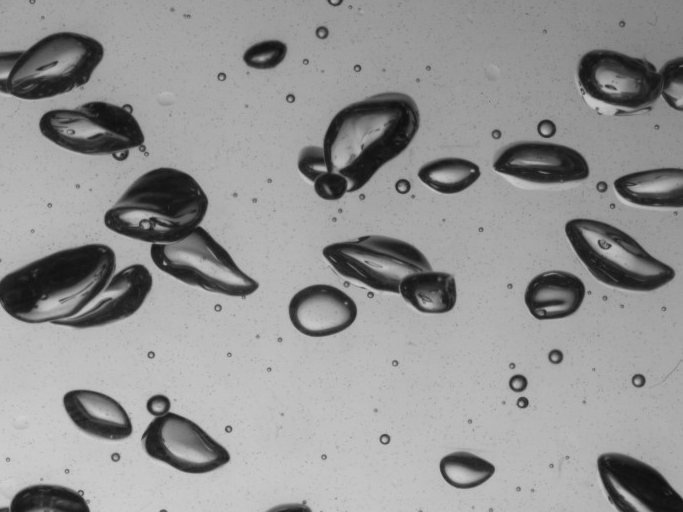}
    }
    \subfloat[][DSB2018 Image]{
    \includegraphics[width=0.15\textwidth]{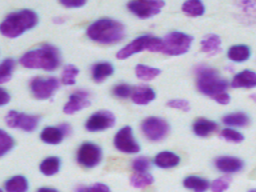}
    }
    \subfloat[][GlaS Image]{
    \includegraphics[width=0.15\textwidth]{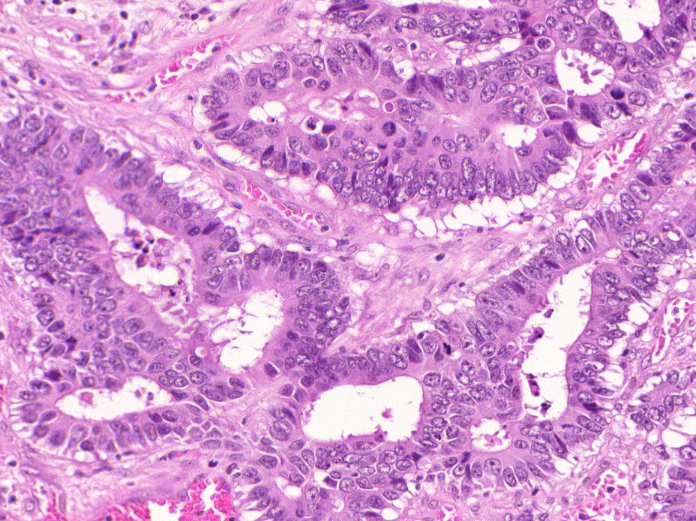}
    }
    
    \subfloat[][Bubble True Mask]{
    \includegraphics[width=0.15\textwidth]{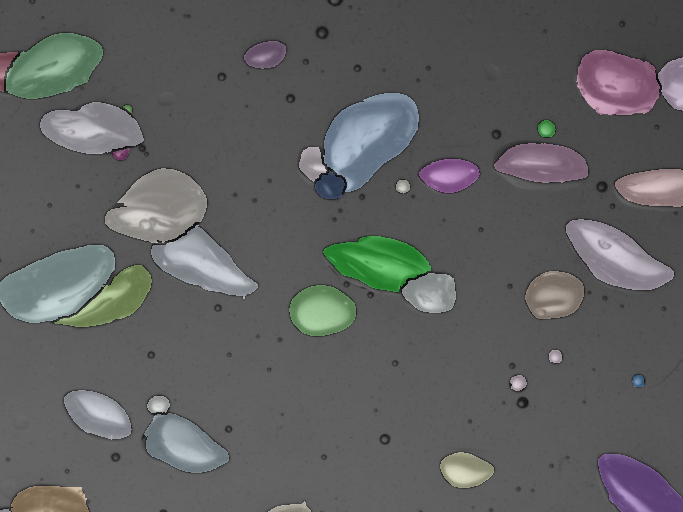}
    }
    \subfloat[][DSB2018 True Mask]{
    \includegraphics[width=0.15\textwidth]{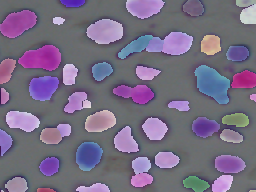}
    }
    \subfloat[][GlaS True Mask]{
    \includegraphics[width=0.15\textwidth]{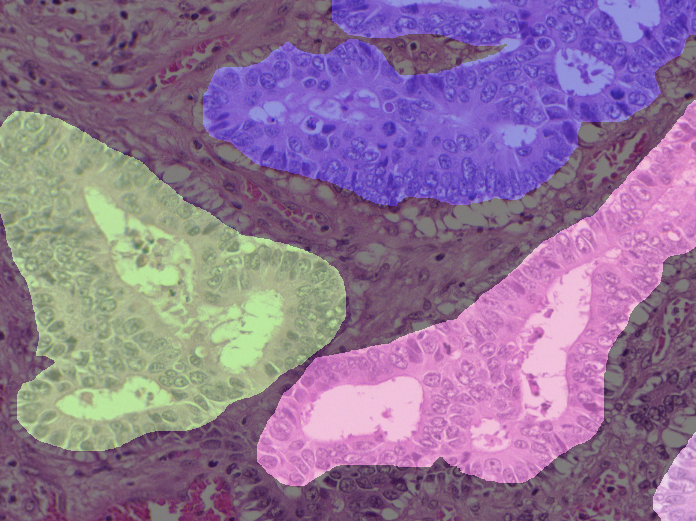}
    }
  
  \caption{Panels (a) to (c) depict the input image for each dataset. Panels (d) to (f) represent the true mask for each dataset.
  }
  \label{fig:datasets}
\end{figure}
\textbf{Bubble:} The Dataset consists of spherical, ellipsoidal, and wobbling bubbles, which are typically encountered in air-water bubbly flows~\cite{bubble}. The dataset includes $414$ manually annotated images of dimensions $256 \times 512 \times 1$, along with corresponding ground truth masks.

\textbf{DSB2018:} Manually annotated real microscopy images of cell nuclei from the 2018 Data Science Bowl~\cite{DSB2018}. The dataset includes $670$ manually annotated images of dimensions $256 \times 256   \times 3$, along with corresponding ground truth masks.

\textbf{GlaS:} Annotated gland segmentation images of Hematoxylin and Eosin stained slides~\cite{GlaS}.The dataset includes $165$ manually annotated images of dimensions $522 \times 775 \times 3$, along with corresponding ground truth masks.

The Figures~\ref{fig:datasets} present examples of input images along with their corresponding true masks for each dataset.
The training set comprises $80\%$ of the images, while the remaining $20\%$ is allocated for validation. We trained our model on the training set and evaluated the estimated certainty on the validation set.

\subsection{Training and Inference}
\textbf{Training:} The StarDist model is employed, featuring three convolutional blocks for down/upsampling. Each block consists of two convolutional layers with $32 \cdot 2^z$ $(z = 0,1,2)$ filters of size $3 \times 3$ and ReLU activation. Following the final upsampling feature layer, an additional convolutional layer with 128 channels and ReLU activation, suggested by Schmidt et al. \cite{stardist_01}, is incorporated. A $n=16$ channel convolutional layer is used to compute the star-convex polygon distance in the StarDist model. The models are trained for 400 epochs.

For the Deep Ensemble technique to estimate certainty, multiple StarDist models are trained using the entire training dataset, with each model randomly initialized. 

In contrast, for the Monte-Carlo Dropout technique to estimate certainty, a single StarDist model is trained with the entire training dataset, supplemented by dropout layers after the last upsampling feature map of the StarDist model. Additionally, in the Monte-Carlo Dropout technique, the impact of the dropout rate on the model's certainty is investigated. The model's certainty is evaluated for dropout rates, $d_{rate} \in \{0.1, 0.5, 0.8\}$.~\footnote{See Supplementary Material for the choice of the three dropout rates.}

\textbf{Inference:} For the Deep Ensemble technique, the validation dataset is used for each of the $F$ StarDist models, and the two approaches (Pixel Approach~\ref{sec:pixel_approach} and the Radial Approach~\ref{sec:radial_approach}) are utilized to cluster and estimate model's certainty. Similarly, in the Monte-Carlo Dropout technique, the validation dataset is processed through a single StarDist model $F$ times to estimate the model's certainty.

%-------------------------------------------------------------------------
\subsection{Quality of Certainty Score}
The calibration diagrams in Figure \ref{fig:cert_quality} with bin size $B = 10$, show the calibration quality of the different certainty scores using the Radial Approach with Monte-Carlo Dropout technique ($d_{rate} = 0.8$, $F = 20$) on the validation set for the Bubble dataset. The diagrams compare hybrid certainty score ($c_{hyb}$), spatial certainty score ($c_{spl}$), and fractional certainty score ($c_{frac}$). 

We observe that the hybrid certainty score $c_{hyb}$ exhibits better calibration as the certainty score closely approximates the expected accuracy (i.e. the bins align to the identity function). The hybrid certainty score considers both spatial similarities and the frequency of instance detections, making it well-suited for more robust comparisons. These comprehensive considerations of factors position the hybrid certainty score as a valuable metric for further comparative analyses.

We observe a consistent pattern on the DSB2018 and GlaS datasets, as illustrated in Figure 1. and Figure 2. of the Supplementary Material.

\subsection{Effects of Forward Passes on Certainties Quality}
\begin{figure*}[t]
  \centering
    \subfloat[][Pearson's R]{
    \includegraphics[width=0.32\textwidth]{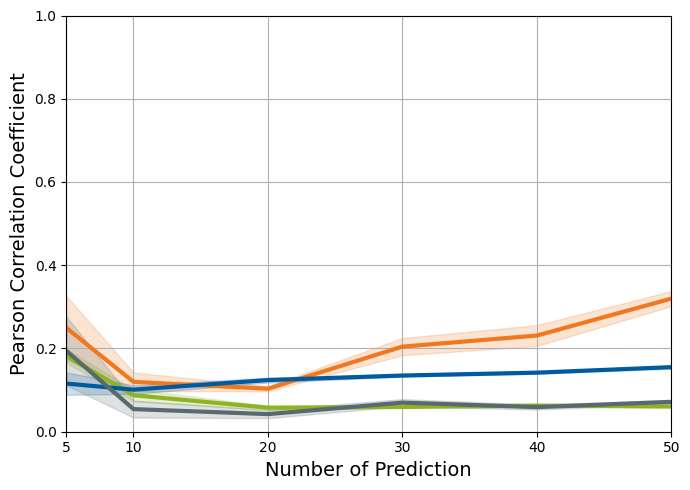}
    \label{fig:fpass_cert_quality_A1_r}}
    \subfloat[][Expected Calibration Error]{
    \includegraphics[width=0.32\textwidth]{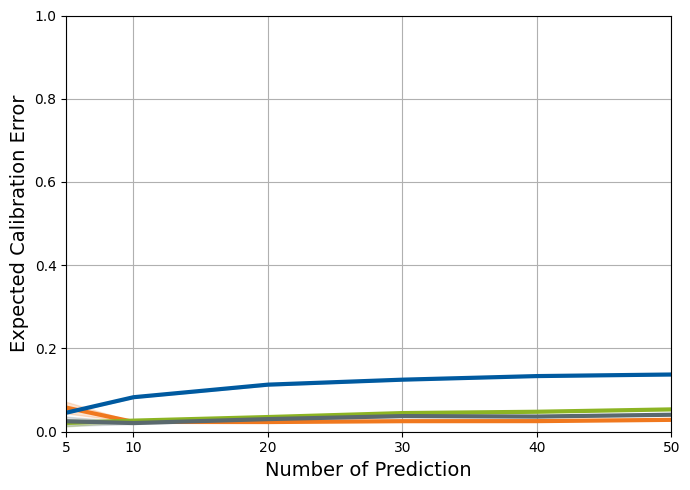}
    \label{fig:fpass_cert_quality_A1_ECE}}
    \subfloat[][Maximum Calibration Error]{
    \includegraphics[width=0.32\textwidth]{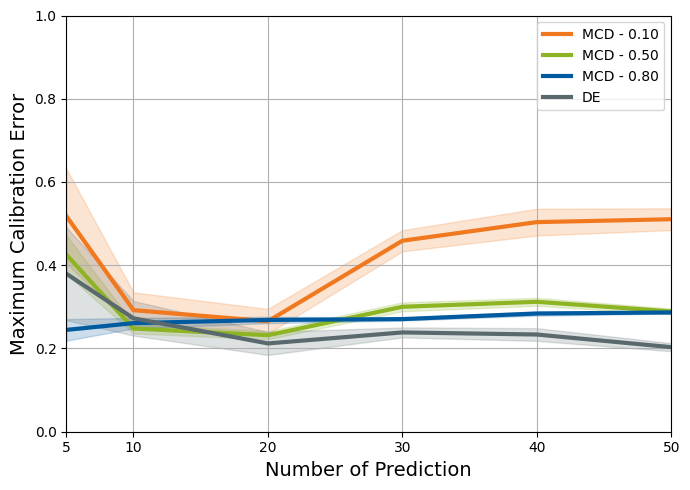}
    \label{fig:fpass_cert_quality_A1_MCE}}
    
    \subfloat[][Pearson's R]{
    \includegraphics[width=0.32\textwidth]{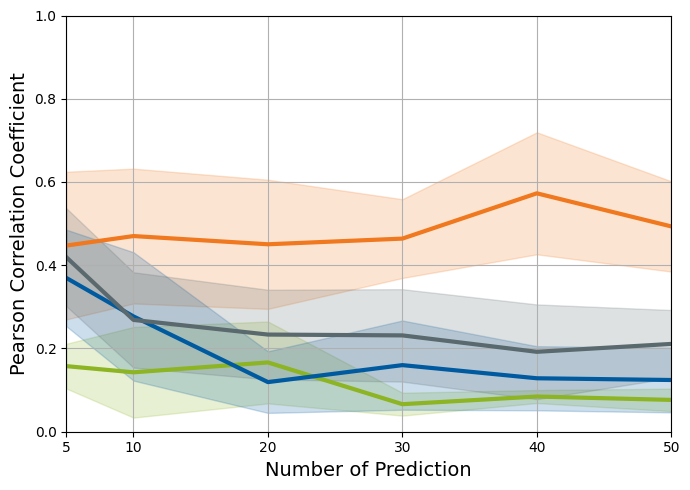}
    \label{fig:fpass_cert_quality_A2_r}}
    \subfloat[][Expected Calibration Error]{
    \includegraphics[width=0.32\textwidth]{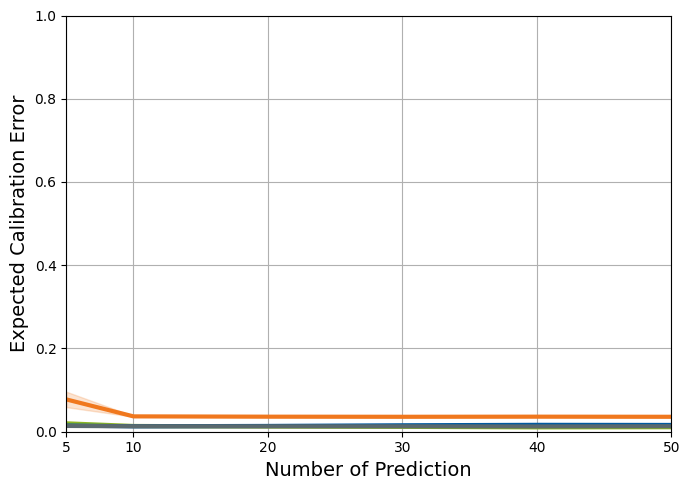}
    \label{fig:fpass_cert_quality_A2_ECE}}
    \subfloat[][Maximum Calibration Error]{
    \includegraphics[width=0.32\textwidth]{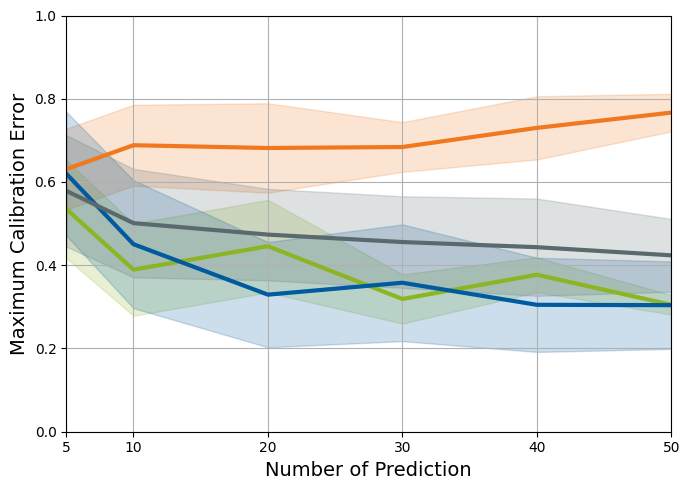}
    \label{fig:fpass_cert_quality_A2_MCE}}
  
  \caption{Plot showing calibration errors as a function of the number of forward passes for the Monte-Carlo Dropout and Deep Ensemble techniques (calibration errors as a function of the number of models in the case of Deep Ensemble). Panels (a) to (c) depict certainty estimates using the Pixel Approach, while panels (d) to (f) represent certainty estimates using the Radial Approach. Notably, the Deep Ensemble technique exhibits faster convergence of calibration errors compared to the Monte-Carlo Dropout technique. Additionally, distinctive convergence patterns are observed for each dropout rate.
  }
  \label{fig:fpass_cert_quality}
\end{figure*}

The influence of the number of forward passes $F$ on the calibration of the hybrid certainty ($c_{hyb}$) was also assessed and the results are visualized in Figure \ref{fig:fpass_cert_quality}. The calibration errors, measured by Pearson's R, Expected Calibration Error (ECE), and Maximum Calibration Error (MCE), are plotted against $F$. 

We observe a convergence of errors as the number of forward passes increases, which aligns with the principles of the Central Limit Theorem.
Additionally, gradually small changes are observed in the error trends after around 20 to 30 forward passes for the Monte-Carlo Dropout technique and 10 models for the Deep Ensemble technique. Furthermore, distinct convergence behaviors are observed for each dropout rate.

We observe consistent behavior when employing both the Pixel Approach and the Radial Approach on both the DSB2018 and GlaS datasets, as depicted in Figure 3. and Figure 4. of the Supplementary Material.

%-------------------------------------------------------------------------
\section{Discussion}
In this work, we introduced and evaluated two approaches to augment instance segmentation predictions by the StarDist model with certainty estimates. The Pixel Approach adapts the work of Morrison \etal~\cite{amazone} to estimate the model's certainty by clustering similar instances based on the $IoU$ score. However, the clustering algorithm used does not scale well as the data increases due to the quadratic complexity.
To address this, the Radial Approach implements a more efficient clustering algorithm leveraging the unique output structure of the StarDist model, scaling the clustering algorithm linearly with the data.

Both Deep Ensemble and Monte-Carlo Dropout techniques yield well-calibrated hybrid certainty estimates, surpassing the spatial and fractional certainty scores individually. The calibration errors decrease for $F = 10$ models in the Deep Ensemble technique, aligning with the findings of Lakshminarayanan et al. \cite{deep_ens} in classification and regression tasks.
%Contrary to expectation, the Monte-Carlo Dropout technique provides the best-calibrated uncertainty estimates.

The results presented in Figure \ref{fig:fpass_cert_quality} further support Lakshminarayanan \etal 's~\cite{deep_ens} claim that the Deep Ensemble technique requires minimal hyperparameter tuning to achieve well-calibrated certainty estimates. They also validate Gal \etal 's~\cite{drop_rate} assertion that the Monte-Carlo Dropout technique's certainty estimates are not calibrated and that the dropout rate must be adjusted to match the model's certainty.
%, typically accomplished through a search for well-calibrated certainty scores. However, conducting a search for large computer vision models can be challenging.

The elevated calibration errors observed in Figure 2. and Figure 4. of the Supplementary Material for the GlaS dataset are due to the dataset's incompatibility with the StarDist model. This leads to higher certainty scores for incorrect predictions, highlighting the critical importance of assessing the calibration of estimated certainties. Such evaluations are pivotal for ensuring model reliability and informed decision-making.

The randomness of the calibration error in Figure 6. Figure 7. and Figure 8. of the Supplementary Material suggests no discernible relationship between the dropout layer's location and its effect on model calibration. Consequently, a search over the locations is necessary for achieving a well-calibrated model.

In conclusion, the most efficient combination for estimating certainty in instance segmentation tasks by StarDist model entails using Deep Ensemble and the Radial Approach, employing $F = 10$ models. This configuration yields well-calibrated results with minimal tuning requirements and an efficient clustering algorithm.

Several possibilities for future expansion of our work exist. One intriguing avenue is exploring the connection between this work and active learning. The obtained certainty information could facilitate the creation of a new dataset containing unknown objects. Ground truth labels for this dataset could be acquired from users, allowing the model's capabilities to continuously adapt to its operating environment through ongoing training.

Additionally, explore the inter-annotator variability as a means to juxtapose the uncertainty inherent in human judgments with the models' epistemic uncertainty. This comparison will shed light on how well the models' uncertainty estimates align with the diversity of human perceptions and annotations, offering deeper insights into the model's capacity to capture uncertainties akin to those observed in human decision-making.

Furthermore, implementing the Concrete Dropout~\cite{con_drop} variant for instance segmentation tasks could be beneficial. This principled extension of dropout enables the tuning of dropout rates, leading to better-calibrated uncertainty estimates in large models while avoiding the coarse and computationally expensive search over dropout rates \cite{drop_rate}.

%-------------------------------------------------------------------------
\section{Summary}
This study addresses certainty estimation in instance segmentation using deep neural networks. While deep neural networks often produce overly confident incorrect predictions, accurate certainty assessment is crucial for informed decisions. Unlike existing methods mainly focused on classification and regression tasks, we pioneer spatial certainty estimation for instances with star-convex shapes.

This study introduces the Radial Approach, a novel clustering method, which combined with Deep Ensemble based certainty sampling provides efficient and well-calibrated certainty estimates.

This research reiterates the significance of calibration for accurate certainty assessment. Calibrating certainty estimates is vital for reliable decision-making and model trustworthiness.

%-------------------------------------------------------------------------

%-------------------------------------------------------------------------

%------------------------------------------------------------------------

%-------------------------------------------------------------------------

%%%%%%%%% REFERENCES
{\small
\bibliographystyle{ieee_fullname}
\bibliography{egbib}
}

\end{document}